\documentclass{article}

\usepackage{arxiv}

\usepackage[utf8]{inputenc} %
\usepackage[T1]{fontenc}    %
\usepackage{hyperref}       %
\usepackage{url}            %
\usepackage{booktabs}       %
\usepackage{amsfonts}       %
\usepackage{nicefrac}       %
\usepackage{microtype}      %
\usepackage{lipsum}
\usepackage{graphicx, caption, subcaption}
\usepackage{caption}
\usepackage{biblatex}
\usepackage{array}
\usepackage{gensymb}
\usepackage[table]{xcolor}

\title{Rotation Invariant Deep CBIR}

\author{
  Subhadip Maji\thanks{GitHub Repo: \emph{https://github.com/pidahbus}} \\
  M.Tech QROR-II \\
  Indian Statistical Institute, Kolkata \\
  Kolkata, 700108 \\
  \texttt{qr1705@isical.ac.in} \\
   \And
 Smarajit Bose \\
  Interdisciplinary Statistical Research Unit \\
  Indian Statistical Institute, Kolkata \\
  Kolkata, 700108 \\
  \texttt{smarajit@isical.ac.in} \\
}
\DeclareUnicodeCharacter{2212}{-}
\begin{document}
\maketitle

\begin{abstract}
Introduction of Convolutional Neural Networks has improved results on almost every image-based problem and Content-Based Image Retrieval is not an exception. But the CNN features, being rotation invariant, creates problems to build a rotation-invariant CBIR system. Though rotation-invariant features can be hand-engineered, the retrieval accuracy is very low because by hand engineering only low-level features can be created, unlike deep learning models that create high-level features along with low-level features. This paper shows a novel method to build a rotational invariant CBIR system by introducing a deep learning orientation angle detection model along with the CBIR feature extraction model. This paper also highlights that this rotation invariant deep CBIR can retrieve images from a large dataset in real-time.
\end{abstract}


\section{Introduction}
\begin{figure}
    \centering
    \includegraphics[scale=0.6]{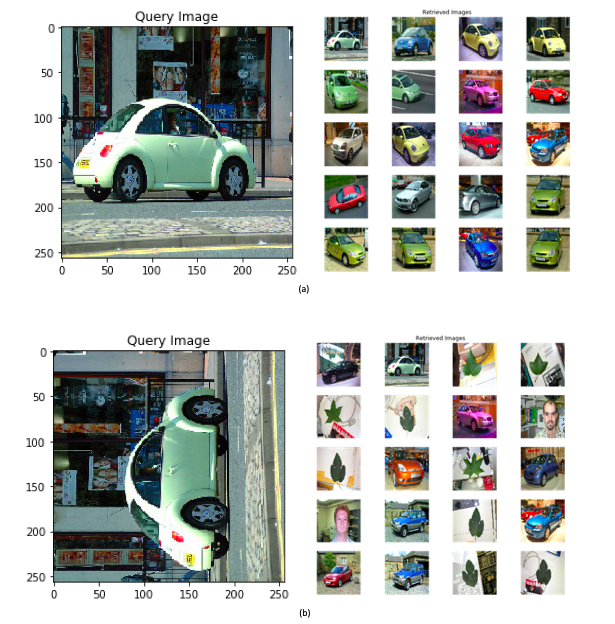}
    \caption{Figure 1: (a) The query image from DB2000\cite{bose} is rotationally correct and results precision of 1. (b) The same query image is rotated 90 degrees anti-clockwise and retrieved almost different images resulting precision value of 0.45. }
    \label{fig:problem_example}
\end{figure}

The pre-trained Convolution Neural Network\cite{maji} for image retrieval is not rotation invariant until now i.e. if we change the orientation of the query image then retrieved images will change. Also, a rotated image results bad retrieval. This can be seen in Figure \ref{fig:problem_example}. Several works are going now to make CNN rotation invariant. One approach is to replace the pooling layer of CNN by a Spatial Transformer Module\cite{jaderberg} which makes it more spatial and eventually CNN becomes rotation invariant. Unfortunately, there is no pre-trained as well as well-defined architecture is not available with the above approach. So, we tried another method to make our CBIR system rotation invariant.

\section{Literature Review}
There is not much work regarding Rotation Invariant CBIR system. However, Bharati et al.\cite{bharati} proposed a method where they analyze curvelet transform and did some useful derivation to extract rotation invariant curvelet features. Sharma et al.\cite{Sharma} introduced a method to extract rotation invariant features using dual tree complex wavelet transform on medical images. Vandhana et al.\cite{raja} described an approach which involves the salient point detection using Scale Up Robust Features (SURF) detector. Milanese et al.\cite{Milanese} presented a method for computing an image signature which is extracted from the Fourier power spectrum by performing a mapping from cartesian to logarithmic-polar coordinates, projecting this mapping onto two 1D signature vectors, and computing their power spectra coefficients. Fountain et al.\cite{fountain} proposed a Rotation Invariant CBIR system by taking the Fourier expansion of the histogram of intensity gradient directions. Tzagkarakis et al.\cite{Tzagkarakis} gave an approach based on a transformation of the texture information via a steerable pyramid to make CBIR rotation invariant. Chifa et al.\cite{chifa} introduced a method which consists of applying circular masks of different size on the image, and extracting the color descriptor from the visible region on the mask, and then combining a texture descriptor for more precision. Krishnamoorthi et al.\cite{Sarfraz} proposed a content-based image retrieval technique with orthogonal polynomials model which extracts texture features that represent the dominant directions, gray level variations and frequency spectrum of the image under analysis and the resultant texture feature vector becomes rotation and scale invariant.

\section{Approach}
\begin{figure}
    \centering
    \includegraphics[scale=0.8]{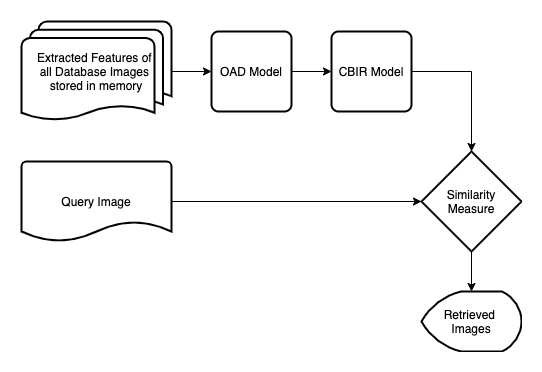}
    \caption{Flowchart of a Rotation Invariant CBIR system}
    \label{fig:flowchart}
\end{figure}

Our idea is to use one pre-trained Orientation Angle Detection Model (OAD) [paper3\_v1] to detect the orientation angle (any angle between 0 to 359 degree) of all the images of our database. Assuming that a given dataset for content-based image retrieval can have any images rotated in any arbitrary angle we will use OAD-360 from [paper3\_v1] and throughout this paper we will term OAD-360 model [paper3\_v1] as only OAD model. Then we will correct the orientation of the database images according to the orientation angle predicted by the model. After that, we will extract features (e.g. 1536 dimensional) feeding the image in our pre-trained CBIR model (e.g. InceptionResNetV2)\cite{maji}. These features we will save in our memory. At the time of retrieval, the query image will pass through both the models: first Orientation Angle Detection Model and second CBIR Model. Then with the extracted features (e.g. 1536 dimensional) from the query image we do similarity measure and retrieve the relevant results. In this way, any image, though fed tilted will be automatically oriented correctly by the OAD model, will retrieve same results if the image was not tilted at all. There may be some slight differences in the results if OAD model fails to predict the correct right orientation angle because of its prediction error. We still tried our best to build our model with minimum validation error. Rotation Invariant CBIR system has been shown in Figure \ref{fig:flowchart}.

\section{Datasets Used}
The CBIR methods were applied on the following image databases, which vary in the number as well as types of images.

\begin{itemize}
    \item \textbf{ImageDB2000:} The database contains 2000 images from 10 different categories. The categories are Flowers, Fruits, Nature, Leaves, Ships, Faces, Fishes, Cars, Animals, and Aeroplanes. Each category consists of 200 images\cite{bose}.
    
    \item \textbf{ImageDBCorel:} This dataset contains 1000 images belonging to 10 categories. Each category contains 100 images. The categories are: African People, Beach, Building, Bus, Dinosaurs, Elephant, Flower, Horse, Mountain and Food\cite{corel}. 
    
    \item \textbf{ImageDBCaltech (Caltech101):} This consists of 9144 images from 102 categories. Each category contains images from 34 to 800 in number\cite{feifei}.
\end{itemize}

\section{OAD Model on CBIR System}
\begin{figure}
    \centering
    \includegraphics[scale=0.8]{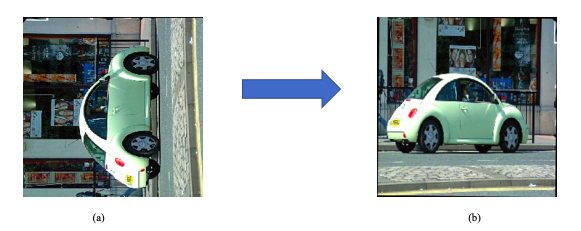}
    \caption{(a) Input Image to OAD Model. (b) OAD Model corrects the orientation angle. }
    \label{fig:oad_correction}
\end{figure}

To tackle the problem showed in Figure \ref{fig:problem_example}, we first passed our rotated image through the OAD model. The OAD model corrects its orientation as shown in Figure \ref{fig:oad_correction}. Now the image after OAD correction (Figure \ref{fig:oad_correction}.b) and Figure \ref{fig:problem_example} are quite similar.

\begin{figure}
  \centering
  \begin{subfigure}{0.45\textwidth}
    \centering
    \includegraphics[scale=0.5]{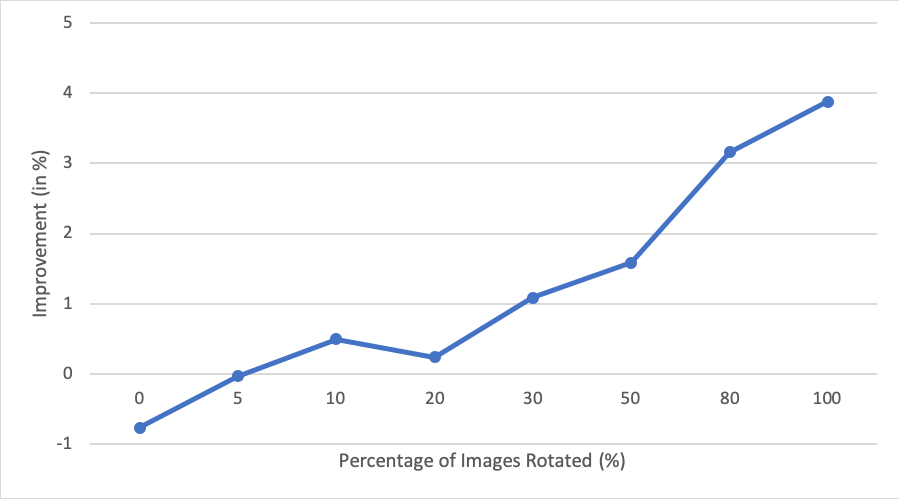}
    \caption{}
  \end{subfigure}
  \begin{subfigure}{0.45\textwidth}
    \centering
    \includegraphics[scale=0.5]{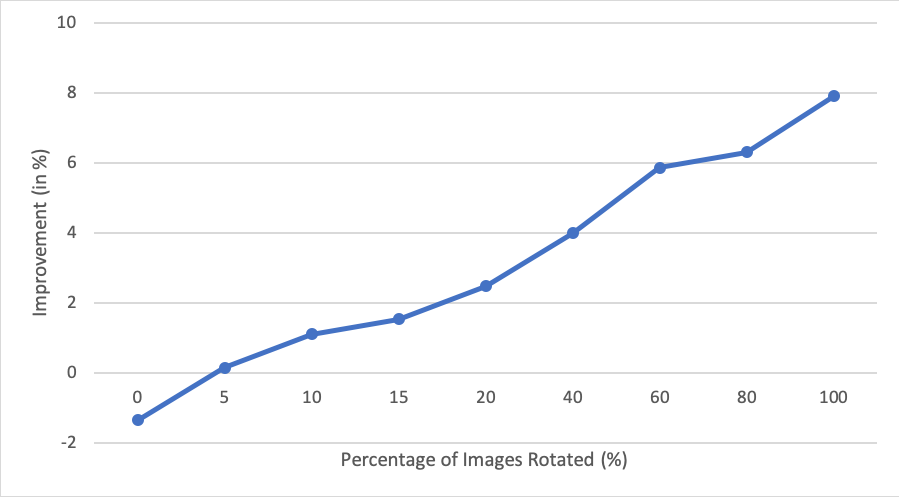}
    \caption{}
  \end{subfigure}
  \caption{Tabulated value of Improvement using OAD model over different values of Percentage of Images Rotated on (a) DB2000 (b) DBCorel. All the values are in \%.}
  \label{fig:oad_graph}
\end{figure}

So, it retrieved the same images improving the precision from 0.45 to 1. To justify the improvement further, we have conducted one experiment on DB2000 and DBCorel Dataset. The idea is to add artificially random rotation of 0 to 359 degrees to the n\% dataset images and extract CBIR features with or without passing the input images through OAD model before feeding into CBIR model, where n=0,5,10,100 etc. Figure \ref{fig:oad_graph} shows the improvement over using OAD model along with CBIR model on DB2000 and DBCorel Dataset. Tabulated result is shown in Table \ref{tab:oad_tab}. Here,

\begin{equation}
    \texttt{Improvement=Precision using OAD Model-Precision without using OAD Model}
\end{equation}

\begin{table}[ht]
    \begin{minipage}[b]{\linewidth}\centering
        \begin{tabular}{|c|c|c|c|}
            \hline
            \textbf{Proportion of Rotation} & \textbf{Before Rotation Model} & \textbf{After Rotation Model} & \textbf{Improvement} \\
            \hline
            100 & 90.275 & 94.155 & 3.88\\
            \hline
            80 & 90.9675 & 94.125 & 3.1575\\
            \hline
            50 & 92.7175 & 94.3025 & 1.585\\
            \hline
            30 & 94.015 & 95.105 & 1.09\\
            \hline
            20 & 94.66 & 94.9 & 0.24\\
            \hline
            10 & 95.075 & 95.575 & 0.5\\
            \hline
            5 & 95.7775 & 95.745 & -0.0325\\
            \hline
            0 & 96.635 & 95.875 & -0.76\\
            \hline
        \end{tabular}
    \caption*{(a)}
    \end{minipage}
    \vspace{0.5cm}
    \begin{minipage}[b]{\linewidth}
    \centering
        \begin{tabular}{|c|c|c|c|}
            \hline
            \textbf{Proportion of Rotation} & \textbf{Before Rotation Model} & \textbf{After Rotation Model} & \textbf{Improvement} \\
            \hline
            100 & 83.6 & 91.5 & 7.9\\
            \hline
            80 & 85.445 & 91.745 & 6.3\\
            \hline
            60 & 86.675 & 92.53 & 5.855\\
            \hline
            40 & 89.135 & 93.13 & 3.995\\
            \hline
            20 & 92.135 & 94.61 & 2.475\\
            \hline
            15 & 92.985 & 94.515 & 1.53\\
            \hline
            10 & 93.5 & 94.605 & 1.105\\
            \hline
            5 & 94.695 & 94.84 & 0.145\\
            \hline
            0 & 96.115 & 94.76 & -1.355\\
            \hline
        \end{tabular}
    \caption*{(b)}
    \end{minipage}
    \caption{Tabulated value of Improvement using OAD model over different values of Percentage of Images Rotated on (a) DB2000 (b) DBCorel. All the values are in \%.}
    \label{tab:oad_tab}
\end{table}

So, positive value of Improvement means Precision increases if we use OAD model along with CBIR Model and negative values tell that precision value decreases if we use OAD model. From the Figure \ref{fig:oad_graph} it is seen that if the dataset has significant number of rotated images, then using OAD model improves the precision of the CBIR system considerably, but if the dataset images are rotationally correct, then using OAD model reduces its precision though a very little amount. We also see that for both the dataset, around 5\% is percentage of rotated images are the number, above which Improvement is positive and below which improvement is negative. So, if any database has around 5-10\% rotated images then we can expect improvement using OAD model along with CBIR model. For large datasets, we can draw a random sample to calculate how many images are rotated.

\section{Real Time Rotation Invariant CBIR}

We have already seen that due to introduction of multiple models i.e. OAD model, CBIR model our results have been improved quite a significant amount, but it arises the retrieval time of images as a matter of question. Whatever models we use, our ultimate goal is to retrieve images in real-time. In this chapter we will discuss about the time complexity of our CBIR system and will show that in spite of introducing several models, our system can retrieve images in real time for DBCaltech and DB2000 dataset.

\subsection{Approach}
Here, we calculate the average query image retrieval time for the scope of 20 on DBCaltech and DB2000 Dataset. This experiment is done with the following combinations:

\begin{itemize}
    \item With OAD Model
    \item Without OAD Model
\end{itemize}

To explain the first combination, at first, we will feed all of our database images through OAD model and CBIR model respectively, then store those extracted features in memory as a feature bank. Now when a query image comes it will be passed through the same OAD model and CBIR model respectively. Then we will compare the extracted features from the query image with each of the feature list in the feature bank and ultimately retrieve those images whose features are closer to the query image features evaluated by some similarity metrics i.e. Manhattan Distance, Euclidean Distance etc. So, the time between the feeding of query image and retrieving similar images is the image retrieval time and Figure 5 shows this average image retrieval time. We use the term “average”, because we used all the images for our database as query image and calculated retrieval time for each of these images and finally took mean. This process is tested on two machines: 

\textbf{Local Machine:}
\begin{itemize}
    \item 1.8GHz Intel Core i5 processor 
    \item 8GB LPDDR3 RAM
\end{itemize}

\textbf{GPU Machine}
\begin{itemize}
    \item GPU: 1 NVIDIA Pascal GPU 
    \item CUDA Cores: 2,048 
    \item Memory Size: 16 GB GDDR5 
    \item H.264 1080p30 streams: 24 
    \item Max vGPU instances: 16 (1 GB Profile) 
    \item vGPU Profiles: 1 GB, 2 GB, 4 GB, 8 GB, 16 GB 
    \item Form Factor: MXM (blade servers) 
    \item Power: 90 W (70 W opt) 
    \item Thermal: Bare Board
\end{itemize}

\begin{figure}
  \centering
  \begin{subfigure}{0.45\textwidth}
    \centering
    \includegraphics[scale=0.8]{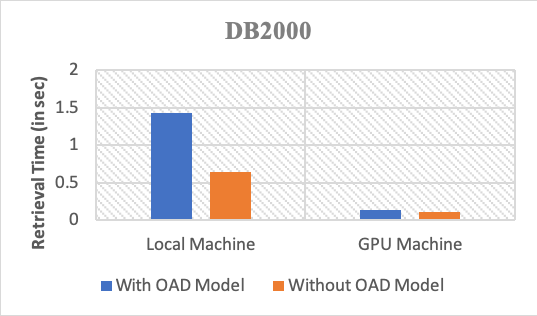}
    \caption{}
  \end{subfigure}
  \begin{subfigure}{0.45\textwidth}
    \centering
    \includegraphics[scale=0.8]{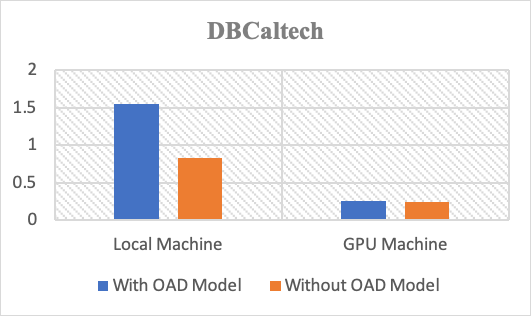}
    \caption{}
  \end{subfigure}
  \caption{Tested image retrieval time on (a) DB2000 (b) DBCaltech}
  \label{fig:time}
\end{figure}

As we all know that GPUs are highly specialized in parallel computing, so the time required for image retrieval is very less in GPU compared to our local machine. This can be clearly seen in Figure \ref{fig:time}. Also from the figure it can be noticed that for GPU machine due to the introduction of OAD model the image retrieval time increases slightly but still it can retrieve image in real time.

DBCaltech has 9144 images with 1536 dimensional features (without PCA) and DB2000 has 2000 images with 1536 dimensional features. We can see that our GPU machine and somewhat our local machine also can retrieve images from these dataset in real time. Image retrieval time depends on both the Database size and dimension. These can be easily verified from Figure \ref{fig:time}. Dimension is almost same as long as we use same architecture (InceptionResNetV2 in our case). But if we use a dataset of very high number of images (say millions) then the image retrieval time increases naturally. In case, where we see that searching through all of the database for relevant images is taking much time then we could use a random sample of size say 10,000 or 20,000 to retrieve 20 images from the database of size millions or billions.  

\section{Conclusion}
This paper shows a novel method to build a rotation invariant CBIR system to handle the CNN features which are not rotation invariant by introducing an intermediate deep learning model to correct the orientation angle of any image. Finally it also shows that introduction of this additional orientation correction model with the existing CBIR model has not have any significant effect in the real time image retrieval, i.e. this rotation invariant deep CBIR can retireve images in real time. 

\clearpage
\printbibliography
\end{document}